\documentclass[review, 1p]{elsarticle}

\usepackage{algorithm}
\usepackage{algpseudocode}
\usepackage{amssymb}
\usepackage{amsfonts}

\usepackage{mypreamble}
\usepackage{setspace}

\def\modelname{RLSyn}
\begin{document}

    \begin{frontmatter}
    
    \title{A Reinforcement Learning Approach to Synthetic Data Generation}

    \author[label1]{Natalia Espinosa-Dice\corref{mycorrespondingauthor}}
    \ead{ne3496@princeton.edu}
    \author[label2]{Nicholas J Jackson}
    \author[label2]{Chao Yan}
    \author[label5]{Aaron Lee}
    \author[label2,label3,label4]{Bradley A Malin}

    \affiliation[label1]{organization={Department of Computer Science, Princeton University},
             city={Princeton},
             state={NJ},
             country={USA}}
    \affiliation[label2]{organization={Department of Biomedical Informatics, Vanderbilt University Medical Center},
             city={Nashville},
             state={TN},
             country={USA}}
    \affiliation[label3]{organization={Department of Biostatistics, Vanderbilt University Medical Center},
             city={Nashville},
             state={TN},
             country={USA}}
    \affiliation[label4]{organization={Department of Computer Science, Vanderbilt University},
             city={Nashville},
             state={TN},
             country={USA}}
    \affiliation[label5]{
        organization={John F. Hardesty Department of Ophthalmology and Visual Sciences, Washington University}, 
        city={St. Louis}, 
        state = {MO}, 
        country={USA}
    }
    \cortext[mycorrespondingauthor]{Corresponding author}
    \tnoteref{codefootnote}
    \fntext[codefootnote]{Code available at \url{https://github.com/natalia-espinosadice/RLSyn}.}

    \begin{abstract}
    \textbf{Objective}:
    Synthetic data generation (SDG) is a promising approach for enabling data sharing in biomedical studies while preserving patient privacy. Yet, state-of-the-art generative models often require large datasets and complex training procedures, limiting their applicability in small-sample settings common in biomedical research. This study aims to develop a more principled and efficient approach to SDG and evaluate its efficacy for biomedical applications.\\

    \noindent\textbf{Methods:} In this work, we reframe SDG as a reinforcement learning (RL) problem and introduce RLSyn, a novel framework that models the data generator as a stochastic policy over patient records and optimizes it using Proximal Policy Optimization with discriminator-derived rewards. We evaluate RLSyn on two biomedical datasets--AI-READI and MIMIC-IV--and benchmark it against state-of-the-art generative adversarial networks (GANs) and diffusion-based methods across extensive privacy, utility, and fidelity evaluations. \\
    
    \noindent\textbf{Results:} 
    On MIMIC-IV, RLSyn achieves predictive utility comparable to diffusion models (S2R AUC 0.902 vs 0.906 respectively) while slightly outperforming them in fidelity (NMI 0.001 vs. 0.003; DWD 2.073 vs. 2.797) and achieving comparable, low privacy risk (~0.50 membership inference risk AUC). On the smaller AI-READI dataset, RLSyn again matches diffusion-based utility (S2R AUC 0.873 vs. 0.871), while achieving higher fidelity (NMI 0.001 vs. 0.002; DWD 13.352 vs. 16.441) and significantly lower vulnerability to membership inference attacks (AUC 0.544 vs. 0.601). Both RLSyn and diffusion-based models substantially outperform GANs across utility and fidelity on both datasets. \\
    
    \noindent\textbf{Conclusion:} Reinforcement learning provides a principled and effective approach for synthetic biomedical data generation, particularly in data-scarce regimes. \\

    \end{abstract}



\end{frontmatter}

    \newcommand{\ned}[1]{\textcolor{blue}{\textbf{[NED #1]}}}
    \newcommand{\nj}[1]{\textcolor{red}{\textbf{[NJ #1]}}}
    
    \doublespacing



    \section{Introduction}\label{sec:introduction}

The amount of biomedical data that is collected through healthcare and research continues to expand at a rapid rate. Unfortunately, our ability to share such data remains limited due to numerous challenges, including concerns over the privacy of patients and research participants \cite{kalkman_patients_2022, schreiber_what_2024, olsen_worldwide_2025}, 
risk aversion \cite{varhol_attitudes_2025, hermansen_heuristics_2023, stieglitz_when_2020}, and general uncertainty regarding regulatory requirements at the local, national, and international level \cite{lencucha_trust_2021, gudi_regulatory_2022} . 
One emerging approach for mitigating these barriers to data sharing is the use of synthetic data. 
\cite{giuffre_harnessing_2023, extance_ai-generated_2025}. Modern synthetic data generation (SDG) methods are typically based on 
generative artificial intelligence (AI) that is trained to create new patient records that emulate the patterns of the original data, thereby preserving key biomedical relationships within the data while severing the link to real individuals \cite{yan_generating_2024, noauthor_synthetic_2025}. 

Despite the promise of generative AI for SDG, state-of-the-art SDG methods have relied on the use of ever larger datasets, which, in turn, have facilitated the use of increasingly complex generative models \cite{kaplan_scaling_2020}. As a result, current state-of-the-art approaches to SDG often leverage
datasets with hundreds of thousands of data points \cite{theodorou_synthesize_2023, yan_multifaceted_2022}, 
limiting their applicability for the smaller datasets that are often used in biomedical research. This trend is driven, in part, by the growing dominance of diffusion-based generative models, which require a computationally demanding denoising process \cite{ho_denoising_2020} that may be infeasible for small datasets. Additionally, prior approaches based on 
generative adversarial networks (GANs) rely on 
adversarial dynamics that are often unstable, data-inefficient, and prone to well-known failures, such as mode collapse.

To address these challenges, we reframe SDG as a reinforcement learning (RL) problem. To the best of our knowledge, the feasibility of RL-based SDG to create private, high-fidelity biomedical datasets has not been assessed. In this formulation, which falls within the class of RL known as inverse-RL, the generative model explicitly defines a parameterized distribution over data records and adjusts it using reward signals that reflect the quality of its generated samples. Furthermore, the generator uses a more principled loss function than other SDG formulations and updates to the generator are bounded through clipped gradients
, encouraging a more stable learning process. We hypothesized that these properties would make RL particularly effective in small-sample settings. 

To test this hypothesis, we introduce \modelname, an RL-based framework for SDG. We then compare \modelname\ to existing GAN-based methods and state-of-the-art diffusion models on two publicly-available medical datasets. We conduct extensive experiments, assessing the utility, fidelity, and privacy of the synthetic data generated by these models. We find that \modelname\  performs comparably to state-of-the-art SDG methods on larger datasets, while outperforming them on smaller datasets, highlighting the utility of \modelname\ for SDG. 



\subsection{Statement of Significance}

\textbf{Problem or Issue:} AI-based generative models are used to create synthetic data that can be broadly shared for scientific use but require large amounts of data to train, making them infeasible for many biomedical applications.

\textbf{What is Already Known:} Synthetic data generation can be used to create biomedical datasets that preserve the statistical properties of the original data, while preserving privacy by severing direct connections to real individuals.

\textbf{What this Paper Adds:} We developed and evaluated \modelname, a novel approach to synthetic data generation based on reinforcement learning that outperforms existing methods, particularly for small datasets.

\textbf{Who Would Benefit:} Scientists seeking to share synthetic data for transparent and reproducible science whose work is based on small, curated datasets.
    
    \section{Related Work}\label{sec:related_work}
We situate our work within prior research on SDG, focusing on generative adversarial networks and diffusion-based approaches. We then provide an overview of reinforcement learning and discuss its prior applications.

\subsection{Generative Adversarial Networks}
GANs are a class of deep learning models that learn to generate realistic data through an adversarial process between two neural networks: a generator that produces synthetic samples and a discriminator that attempts to distinguish between real and fake data \cite{goodfellow_generative_2014}. Through this competition, the generator learns to produce data that closely approximates the true distribution of the input data. The use of GANs for SDG in healthcare was first identified by Choi et al. with the invention of medGAN \cite{choi_generating_2017}. Since then, GANs have been studied for their ability to generate many different biomedical modalities, including imaging \cite{jackson_enhancement_2024, frid-adar_gan-based_2018} and text in the form of clinical notes \cite{baumel_controllable_2024, li_are_2021}. Additionally, much work has gone into improving GANs for tabular data generation \cite{yan_multifaceted_2022, zhang_ensuring_2020, baowaly_synthesizing_2019}. The primary drawback of GANs is that it can be difficult to learn an effective discriminator. For example, if the discriminator is too strong and easily distinguishes real from synthetic samples, then the gradient updates 
become negligible and the generator does not improve. On the other hand, if the discriminator is too weak (i.e., it 
cannot effectively distinguish real from synthetic data), then it 
will fail to provide meaningful feedback to the generator \cite{arjovsky_towards_2017}. These issues have been partially addressed by modifying the GAN loss function \cite{arjovsky_wasserstein_2017}; however, instabilities in GAN training persist due to their sensitivity to hyperparameters and the non-stationarity of the loss function \cite{lin_why_2021}.

\subsection{Denoising Diffusion Models}

Denoising diffusion models \cite{ho_denoising_2020} are a class of generative models that synthesize data by learning to reverse a gradual noising process, in which noise is incrementally added to real data and a neural network is trained to predict and remove this noise at each step. At generation time, the model starts from random noise and iteratively denoises it to produce realistic samples. These models have exhibited state-of-the-art performance in generating high-quality synthetic images \cite{ho_denoising_2020, rombach_high-resolution_2022}. For tabular healthcare data, several diffusion-based approaches have been developed, appearing to outperform their GAN-based counterparts \cite{yuan_ehrdiff_2024, yoon_ehr-safe_2023}. While diffusion models are considered more principled approaches to SDG than GANs, they suffer from two non-trivial deficiencies in the context of SDG for healthcare. First, diffusion models have been shown to memorize training data, occasionally reproducing nearly identical copies, which may limit their applications to data sharing in healthcare as this may pose a risk to data privacy \cite{carlini_extracting_2023, rahman_investigating_2025}. Second, diffusion models require large amounts of training data, which may make them difficult to train from scratch for healthcare applications.

\subsection{Reinforcement Learning}
RL is a framework for sequential decision-making in which an agent learns to take actions in an environment to maximize cumulative reward. Unlike supervised learning, which relies on labeled examples, RL learns through interaction: the agent observes the current state, selects an action, receives feedback in the form of a reward and updates its behavior based on this experience. Over time, the agent learns a policy (i.e., a mapping from states to actions) that aims to achieve the highest long-term return. RL has been applied across a wide range of domains that involve complex, sequential decisions, including robotics and games such as Go, chess and Atari \cite{schrittwieser_mastering_2020}. In healthcare, RL has been explored as a tool for optimizing sequential clinical decisions and treatment strategies \cite{komorowski_artificial_2018}. RL has also been incorporated into generative modeling, although to a lesser extent. Most notably, SeqGAN \cite{yu_seqgan_2017} and ORGAN \cite{guimaraes_objective-reinforced_2018} use RL to guide sequence generation of text and musical melodies using a GAN backbone. However, these approaches rely on high variance policy-gradient updates and are designed specifically for discrete sequential data. To our knowledge, RL has not been applied to generating high-dimensional clinical tabular data or explored as a general framework for synthetic data generation.

    \section{Methodology}\label{sec:methodology}
\subsection{Datasets}
In this paper, we utilized two publicly-available medical datasets: 1) Artificial Intelligence Ready and Exploratory Atlas for Diabetes Insights (AI-READI) \cite{ai-readi_consortium_ai-readi_2024} and 2) Medical Information Mart for Intensive Care (MIMIC)-IV \cite{johnson_mimic-iv_2023}. 

The first dataset is based on AI-READI version 2.0, which contains data on 1067 patients spanning the full spectrum of type 2 diabetes severity, collected across multiple sites in the United States between 2023 and 2024. We extracted tabular data on participant diagnoses and laboratory measurements, as well as automatic measurements from electrocardiograms. Additionally, participants in AI-READI spent several days wearing wearable activity monitors and continuous glucose monitors. 
We processed the wearable time-series data by aligning measurements from multiple sensors onto a common daily time grid with 5-minute resolution, anchored at 08:00 AM. Short gaps due to irregular sampling were interpolated, with interpolation restricted to gaps of at most three hours. Days with insufficient sensor coverage, extended gaps beyond this threshold, or physiologically implausible values (e.g., calorie expenditure exceeding 20 kcal/min) were excluded. We then split the remaining data by person-days and computed the number of hours in each of the four sleep stages and the four activity stages. We computed the mean heart rate and respiratory rate over the course of the day as well as the mean and standard deviation of their blood glucose. We computed participants' resting heart rate during sedentary periods and their total calories burned and steps over each day. Lastly, we computed the mean of a stress score provided by the fitness wearable. We concatenated the 16 wearable-derived features with 56 continuous features consisting of clinical measurements (e.g., insulin levels) and measures derived from electrocardiograms (e.g., QT duration). We combined the resulting 72 continuous features with the most commonly-recorded 36 clinical conditions, yielding a dataset of 4,537 person-days (rows) of tabular data, each with 108 features. A complete list of features can be found in Appendix Table \ref{tab:aireadifeature_list}. 

The second dataset is based on MIMIC-IV version 2.0, which contains data on over 300,000 patients who visited the emergency department at Beth Israel Deaconess Medical Center in Boston, Massachusets. We preprocessed MIMIC-IV as in Yan et al. \cite{yan_generating_2024}, such that we included only patients who were admitted to the hospital and did not die during their visit. We then extracted each patient's demographic information (age, race, and gender), all of their documented diagnoses, BMI, and blood pressure. Additionally, we mapped diagnoses from International Classification of Disease (ICD) codes to clinically-meaningful phenotypes using phecodes \cite{bastarache_using_2021}. This yielded 180,746 patient-stays, each with 1,486 features.

\subsection{\modelname}

In RL-SYN, the generator is trained as a \emph{policy}, a probabilistic model that defines a distribution over full patient records. Rather than outputting a single deterministic sample, the generator specifies a probability for each feature value. Sampling from this distribution yields a complete synthetic record. The training goal is to adjust this distribution so that it generates data that closely resemble real patient records. 

\begin{figure}[!htb]
\centering
\includegraphics[width=0.5\textwidth]{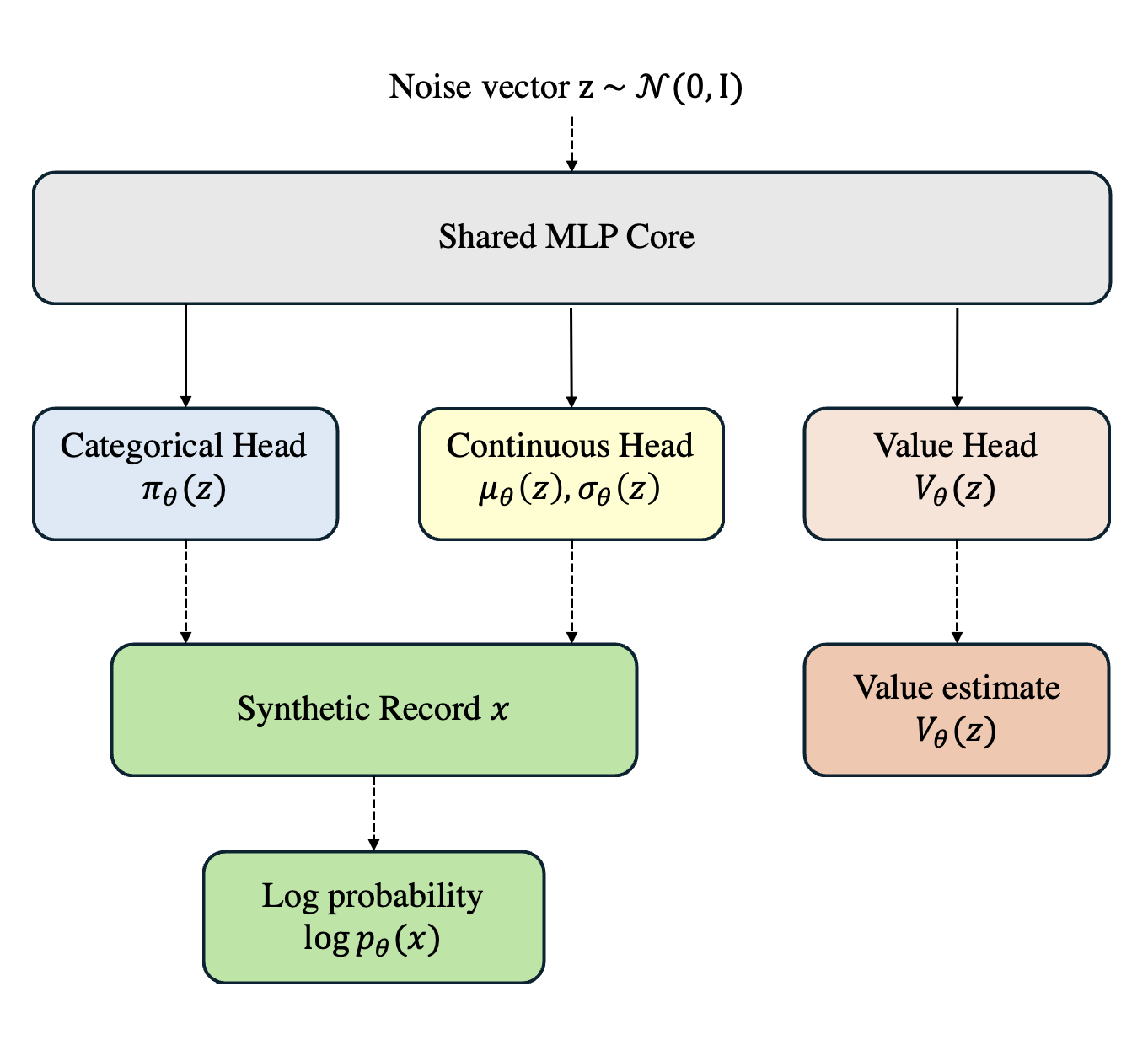}
\caption{RL-SYN Generator Architecture. A shared multilayer perceptron (MLP) maps a latent noise vector to three output heads: a categorical head that parameterizes a distribution over discrete features, a continuous head that parameterizes a distribution over continuous features, and a value head that estimates the expected reward of the generated record. The categorical and continuous outputs are combined to form a synthetic record $x$, whose log-probability under the generator and associated value estimate are used to update the generator during training.}
\label{fig:genarch}
\end{figure}

In practice, the policy distribution is implemented as a combination of continuous and categorical components. For continuous features, the generator outputs the mean and variance that parameterize a Gaussian distribution, from which samples are drawn, passed through a $\tanh$ transformation and constrained to the range $(0, 1)$. For categorical and binary features, the generator outputs logits that parameterize independent Bernoulli distributions. Together, these define a joint distribution $p_\theta(x)$ over all feature values, representing the generator's probabilistic policy. The generator also maintains a value baseline that is utilized during policy optimization. Figure \ref{fig:genarch} illustrates the generator architecture. 

The learning process follows a reinforcement learning loop in which the policy is refined based on feedback from the discriminator. The discriminator evaluates each generated record and provides a \emph{reward} indicating how realistic it appears. The generator then updates its parameters to increase the likelihood of producing samples that yield higher rewards. Although this reward is derived from a discriminator, 
the generator is not updated through adversarial gradient descent as in GANs. Rather, the discriminator serves only as a reward estimator and the generator is optimized independently through its own RL objective with no discriminator gradients passing through. This avoids the minimax component of GAN training and yields a more stable update process. 

The generator update is implemented using Proximal Policy Optimization (PPO), a stable and widely used policy-gradient algorithm \cite{DBLP:journals/corr/SchulmanWDRK17}, as shown in Algorithm \ref{alg:ppo}. 
At each iteration, the discriminator provides a reward for each generated sample and the generator computes an advantage comparing this reward to its own value baseline (Line~2). This advantage is normalized to stabilize learning (Line~3). To encourage the continuous features to remain close to the real data distribution, we include a light mean-matching penalty (Line~4). During each PPO update epoch (Lines~5-12), the generator recomputes log-probabilities and value estimates under the current policy (Line~6). To ensure stable learning, PPO computes the probability ratio between the updated and previous policies (Line~7) and applies a clipped surrogate objective (Line~8), which prevents the generator from making large, destabilizing shifts in probability mass while still encouraging actions that yield higher rewards. The value head is trained to predict observed rewards (Line~9), and an entropy term encourages sample diversity (Line~10). All components are combined into the final generator loss (Line~11), which is then used to update the generator's parameters (Line~12).

After updating the generator, we next update the discriminator to maintain an accurate reward signal. During each discriminator step (Algorithm \ref{alg:rlsyn}, Lines 9–14), the discriminator is trained to distinguish real records from synthetic records using a standard binary cross-entropy loss. A small R1 gradient penalty is applied on real samples to improve smoothness and prevent overfitting. 

\begin{algorithm}[!htb]
\caption{Generator PPO Update}
\label{alg:ppo}
\begin{algorithmic}[1]
\Require{Real data $X_{\text{real}}$; PPO clip $\epsilon$; entropy weight $\beta$; value weight $c_v$; mean penalty $\lambda_m$; learning rate $\eta_G$.}
\State Evaluate reward $r = \sigma(D_\phi(x))$ 
\State Compute advantage $\hat{A} = r - V_\theta(z)$ 
\State Normalize advantage $\hat{A} \leftarrow (\hat{A} - \mathrm{mean}(\hat{A}))/(\mathrm{std}(\hat{A}) + 10^{-8})$
\State Compute mean-matching term $M = \|\text{mean}(x^{(c)}) - \text{mean}(X_{\text{real}}^{(c)})\|_2^2$ 
\For{PPO update epochs}
\State Recompute $\log p_\theta(x)$ and $V_\theta(z)$ under current generator policy 
\State Compute probability ratio $\rho = \exp(\log p_\theta(x) - \log p_{\theta_{\text{old}}}(x))$ 
\State Compute clipped objective $L_{\text{clip}} = \mathbb{E}[\min(\rho \hat{A}, \text{clip}(\rho,1-\epsilon,1+\epsilon)\hat{A})]$ 
\State Compute value baseline $L_V = \mathbb{E}[(V_\theta(z) - r)^2]$ 
\State Compute entropy term $H = -\mathbb{E}[\log p_\theta(x)]$ 
\State $\mathcal{L}_G = -L_{\text{clip}} + c_v L_V - \beta H + \lambda_m M$
\State $\theta \leftarrow \theta - \eta_G \nabla_\theta \mathcal{L}_G$
\EndFor
\end{algorithmic}
\end{algorithm}

\begin{algorithm}[!htb]
\caption{RL-Syn}
\label{alg:rlsyn}
\begin{algorithmic}[1]
\Require Real data $X_{\text{real}}$; noise dimension $k$; learning rate $ \eta_D$;  gradient penalty $\gamma$; discriminator steps $K$.
\For{iteration $t = 1, \dots, T$}  
    \State $z \sim \mathcal{N}(0, I_k)$
    \State $x, \log p_{\text{old}}, V_{\theta_\text{old}} = G_\theta.\text{sample}(z)$ 
    \State Run \textbf{Algorithm 1} (Generator PPO Update)
    \For{$k = 1, \dots, K$} \Comment{discriminator update}
        \State Sample real and synthetic $x_r \sim X_{\text{real}}$, $x_f \sim G_\theta$
        \State $L_{\text{BCE}} = \text{BCE}(D_\phi(x_r),1) + \text{BCE}(D_\phi(x_f),0)$
        \State $R1 = \frac{\gamma}{2}\mathbb{E}[\|\nabla_{x_r} D_\phi(x_r)\|_2^2]$
        \State $\mathcal{L}_D = L_{\text{BCE}} + R1$
        \State $\phi \leftarrow \phi - \eta_D \nabla_\phi \mathcal{L}_D$
    \EndFor
\EndFor
\State \Return $G_\theta$
\end{algorithmic}
\end{algorithm}

\subsection{Baselines}
We compared \modelname\ against two baseline generative models developed specifically to promote privacy in the sharing of tabular health data: Electronic Medical Record Wasserstein GAN (EMR-WGAN) \cite{zhang_ensuring_2020} and Electronic Health Record Diffusion (EHRDiff) \cite{yuan_ehrdiff_2024}. EMR-WGAN was specifically designed for its use in structured EHR data and has shown strong performance across multiple datasets, outperforming other GAN architectures \cite{yan_multifaceted_2022}. EHRDiff is a diffusion model that, like EMR-WGAN, was developed specifically for structured EHR data. It has been shown to outperform EMR-WGAN on specific EHR data generation tasks, while maintaining privacy. Collectively, these models represent existing state-of-the-art algorithms for tabular health record generation. 

\subsection{Evaluation Criteria}

\subsubsection{\textbf{Privacy}}

We evaluated the privacy of the synthetic data generated by each model in terms of membership inference risk. In our setting, membership inference occurs when a simulated adversary with access to an individual's health information, correctly infers that the individual's data was 
involved in training the generative model. 
We selected real patient records that were in the training set 
for the generative model and records that were in the held-out test set. We then calculated the Euclidean distance between each real record and the closest synthetic record (with features normalized to a range of [0,1]). We 
claim that the real record is in the synthetic data if this distance is less than a predefined threshold. We then compute the sensitivity and specificity.
We perform this analysis over a range of thresholds to construct a receiver operatic characteristic (ROC) curve and compute the corresponding AUROC.
For AI-READI, we used all training and testing data to perform this attack; however, it was computationally 
infeasible to run this analysis on the entirety of the MIMIC-IV dataset. As such, for the latter, we relied upon a subsample of 30,000 records (following a 70/30 train/test split). 
To verify the efficacy of this attack, we performed it against the real data as well.

\subsubsection{\textbf{Utility}}

To assess data utility, we investigated how useful the synthetic data is for a set of classification tasks. For AI-READI, we define this task as the training of machine learning models to classify type 2 diabetes. For MIMIC-IV, this task is the prediction of mortality within 1 year. We first established the baseline performance of machine learning models for these tasks on the real data. We then trained models on the synthetic data and tested them on the real testing data, a setting we refer to as synthetic-to-real or S2R. These analyses allow us to evaluate whether the synthetic data sufficiently captures the key statistical properties and relationships present in the real data. Specifically, if the synthetic data supports comparable model performance to real data, this suggests that it successfully represents the underlying patterns in the original dataset. To ensure fair comparisons for these analyses, we held the size of the synthetic training and testing datasets to be the same as their respective real counterparts.

\subsubsection{\textbf{Fidelity}}

\textit{Real-to-Synthetic Classification}

Similar to the utility evaluations, we trained models on the real training data and tested them on the synthetic data, a setting we refer to as real-to-synthetic or R2S. As with S2R, if R2S classification performance matches that of models trained and tested on the real data, this indicates that the same statistical relationships in the real data are present in the synthetic data. While some prior work has considered R2S to be a utility metric, we classify it as a fidelity measure. This is because R2S values that are higher than performance on the real data do not indicate higher quality synthetic data. A key detail here is that the R2S analysis trains the classifier on the same real data that the synthetic data was generated from. Consequently, if the R2S results are higher than the R2R baseline, this indicates that the generative model may have simplified the synthetic data by removing outliers, over-fitting, or replicating close proxies of the real data. As a result we aim for S2R classification performance that matches, but does not exceed the real data counterpart.

\textit{Latent Space Comparison}

This metric evaluates the difference between a synthetic dataset and its corresponding real dataset in the latent space using unsupervised clustering \cite{yan_multifaceted_2022}. We combined each pair of real and synthetic datasets into a single
dataset and applied principal component analysis (PCA) to reduce dimensionality, retaining the components that account for 80\% of the variance. Following this, we performed $k$-means clustering to identify clusters and selected the optimal number of clusters using the elbow method \cite{yan_generating_2024}
. Subsequently, we computed the Normalized Mutual Information (NMI) 
between the cluster assignments and whether the synthetic data was real vs synthetic. This metric ranges from 0 to 1 with a lower value indicating that the distributions of real and synthetic data are more similar (i.e., there is no relationship between which cluster the data were assigned to and whether the data was synthetic or real).

\textit{Column-wise Correlation Differences (CWC)}

This metric captures the extent to which the synthetic data retain the correlations present in the real data. We computed linear (Pearson) correlations between all features in the real and synthetic datasets, separately. We then computed the average absolute difference between the real and synthetic correlation matrices. A lower value of for this metric indicates better preservation of the correlations in the original data.

\textit{Dimension-wise Differences}

Dimension-Wise Difference (DWD) is an aggregate measure of how well individual dimensions (i.e., features) are represented in the synthetic data. DWD is assessed differently for categorical and continuous features.
For categorical features (i.e., diagnoses) we computed the absolute prevalence difference (APD) between the real and synthetic datasets. Similarly, we computed the average Wasserstein distances (AWD) for continuous features (e.g., BMI, blood pressure). After normalizing the AWDs to $[0,1]$, DWD is computed as the sum of the AWDs and APDs for each feature. A lower value of this Dimension-Wise Difference (DWD) score indicates higher similarity between the real and synthetic data along individual dimensions. We visualize the APDs through bigram plots, assessing the prevalence of all categorical features in the real dataset as compared to those in the synthetic dataset. Similarly, we selected a subset of the continuous variables and visualized their distributional differences using histograms. 

\subsection{Model Training}

We partitioned AI-READI and MIMIC-IV into 90-10 and 70-30 training and testing datasets, respectively. To ensure the robustness of our analysis, we tuned the hyperparameters of each SDG model-dataset pair for 20 trials using Optuna \cite{10.1145/3292500.3330701}, selecting the hyperparameters that maximized classification performance for the S2R utility analysis. However, for EMR-WGAN on MIMIC-IV, we did not perform additional tuning, as we used the exact code and hyperparameters as described in \cite{yan_generating_2024}. The final hyperparameters chosen are provided in Appendix Table A.3. After selecting the optimal hyperparameters, we trained each SDG model 10 times for each dataset across 10 different train-test splits and averaged all performance measures accordingly.

    \section{Results}\label{sec:results}

\begin{table}[h!]
    \centering
    \caption{Area under the receiver operating characteristic curve (AUC) of membership inference attacks against synthetic data.}
    \begin{tabular}{c|c|c}
        \toprule
          & \textbf{MIMIC-IV} & \textbf{AI-READI} \\
        \cmidrule{1-3}
        Real Data & 0.998 $\pm$ 0.000 & 1.000 $\pm$ 0.000 \\
        \cmidrule{1-3}
        EMR-WGAN & 0.498 $\pm$ 0.002 & 0.501 $\pm$ 0.026\\
        EHRDiff & 0.499 $\pm$ 0.002 & 0.601 $\pm$ 0.036\\
        \modelname & 0.499 $\pm$ 0.002 & 0.544 $\pm$ 0.030\\
        \bottomrule
    \end{tabular}
    \label{tab:grouped}
\end{table}

\subsection{Privacy}
Table 1 summarizes the results of the membership inference attacks for the various SDG models. For MIMIC-IV, all three models performed 
admirably, generating synthetic data for which membership inference attacks was no better than random chance (AUC $\approx 0.5$). However, the attack exhibited greater success against AI-READI. 
While EMR-WGAN was resilient to the attack,
with a membership inference AUC of 0.501, \modelname\ was slightly less private with an AUC of 0.544, and EHRDiff exhibited an AUC of 0.601, which was well above random chance. 

\begin{figure}[!htb]
\centering
\includegraphics[width=1.0\textwidth]{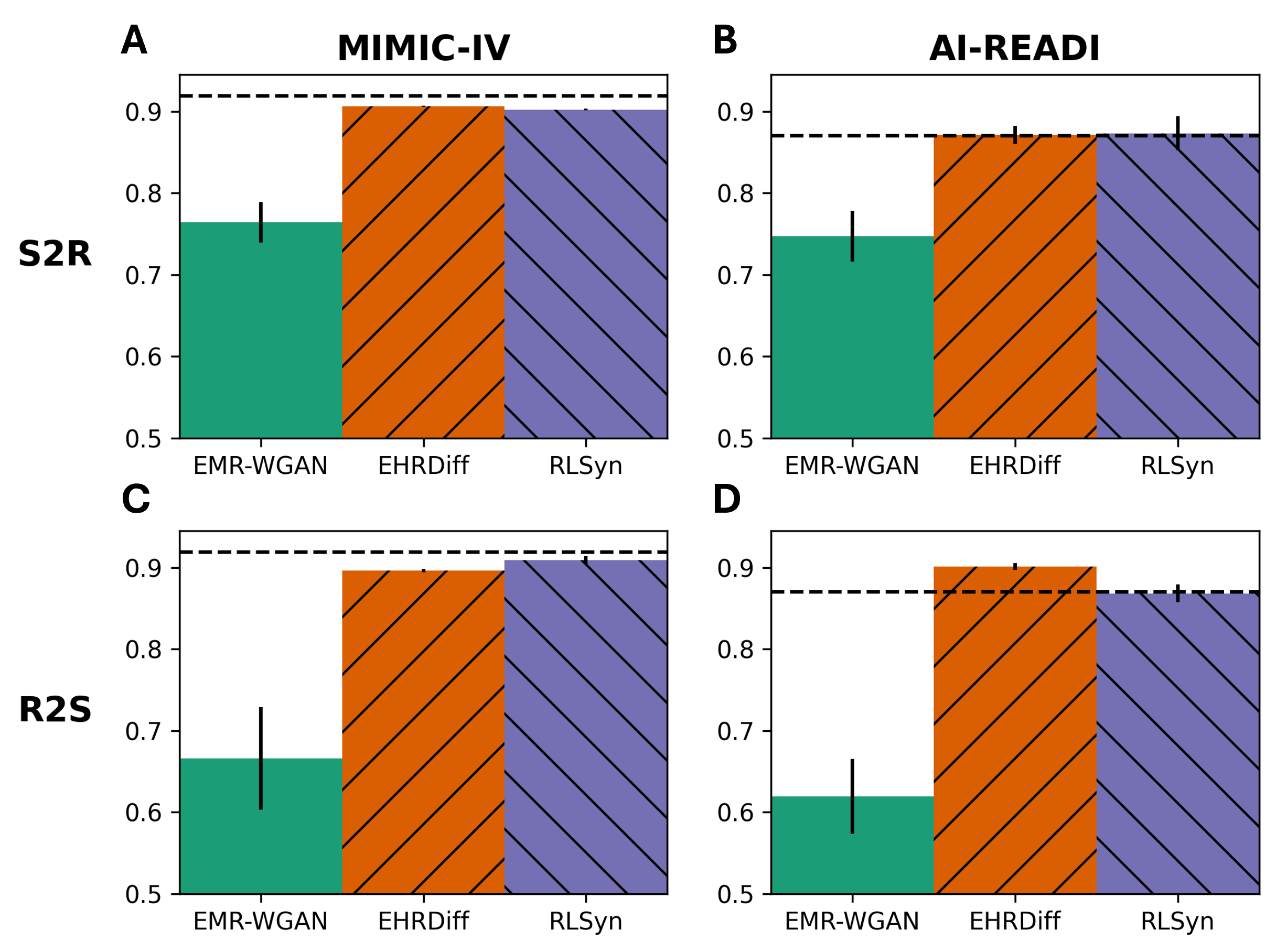}
\caption{A comparison of the performance of classifiers trained on synthetic data and tested on the real test dataset (S2R) and vice versa (R2S) for both the MIMIC-IV and AI-READI datasets. The dashed black line represents the baseline of classifiers trained and tested on real data (R2R). Error bars represent 95\% confidence intervals.}
\end{figure}

\subsection{Utility}
Figure 2, row 1 summarizes the utility results.
The S2R analysis on the MIMIC-IV dataset indicated that EHRDiff and \modelname\ were the best-performing with respectable classification AUCs of 0.906 and 0.902, respectively. EMR-WGAN trailed with an AUC of 0.764. Similar results were observed in the AI-READI dataset. Specifically, \modelname\ and EHRDiff had comparable S2R performance (AUC: 0.873 and 0.871, respectively), both outperforming EMR-WGAN (AUC: 0.747).

\begin{table}[h!]
    \centering
    \scriptsize
    \setlength{\tabcolsep}{2pt}

    \caption{Fidelity results measured via column-wise correlation difference (CWC), Normalized Mutual Information (NMI), and Dimension-Wise Difference (DWD). For all three metrics lower scores represent higher fidelity. The best performance for each metric is bolded.}
    \begin{tabular}{c|c|c|c|c|c|c}
        \toprule
        \multicolumn{1}{c|}{} & 
        \multicolumn{3}{c|}{\textbf{MIMIC-IV}} & 
        \multicolumn{3}{c}{\textbf{AI-READI}} \\
        \cmidrule{1-7}
        
          & \textbf{CWC} ($\downarrow$) & \textbf{NMI} ($\downarrow$) & 
          \textbf{DWD} ($\downarrow$) & 
          \textbf{CWC} ($\downarrow$) & \textbf{NMI} ($\downarrow$) &
          \textbf{DWD} ($\downarrow$)\\
        \cmidrule{1-7}
        EMR-WGAN & 11.852 $\pm$ 0.045 & 0.320 $\pm$ 0.130 & 17.642 $\pm$ 0.602 & 1.352 $\pm$ 0.055 & 0.809 $\pm$ 0.003 & 77.056 $\pm$ 0.227\\
        EHRDiff & \textbf{7.848 $\pm$ 0.033} & 0.003 $\pm$ 0.000 & 2.797 $\pm$ 0.045 & \textbf{0.427 $\pm$ 0.007} & 0.002 $\pm$ 0.000 & 16.441 $\pm$ 10.989\\
        \modelname & 8.877 $\pm$ 0.200 & \textbf{0.001 $\pm$ 0.000} & \textbf{2.073 $\pm$ 0.073} & 0.750 $\pm$ 0.0015 & \textbf{0.001 $\pm$ 0.001} & \textbf{13.352 $\pm$ 8.325} \\
        \bottomrule
    \end{tabular}
    \label{tab:grouped}
\end{table}

\subsection{Fidelity}

Figure 2 
details the results of the R2S experiments, the results of which were similar to those of 
the S2R experiments. Overall, EMR-WGAN performed poorly on the MIMIC-IV and AI-READI when compared to EHRDiff and \modelname. While \modelname\ outperformed EHRDiff on MIMIC-IV (0.909 vs 0.896), the opposite was observed for AI-READI (0.868 vs 0.901). Notably, EHRDiff outperformed the R2R baseline of 0.870 on 
AI-READI.

For the latent space comparison, clustering was performed using the elbow method, resulting in 3 clusters for AI-READI and 8 clusters for MIMIC-IV.
The results for the other fidelity metrics mirror those of the R2S analysis in that \modelname\ had the lowest NMI between clusters and CWC difference, outperforming both EMR-WGAN and EHRDiff for AI-READI (Table 2). Similarly, for MIMIC-IV \modelname\ and EHRDiff were similar in terms of NMI (0.001 vs 0.003) and CWC (8.877 vs 7.848). Notably, both sets of values were much lower than their counterparts generated by EMR-WGAN (NMI: 0.320, CWC: 11.852). These differences are made visually apparent in Figure 3, which plots the principal components of the real and synthetic datasets. It can be seen that that \modelname\ more closely matches the real data distribution for both datasets.

\begin{figure}[!htb]
\centering
\includegraphics[width=1.0\textwidth]{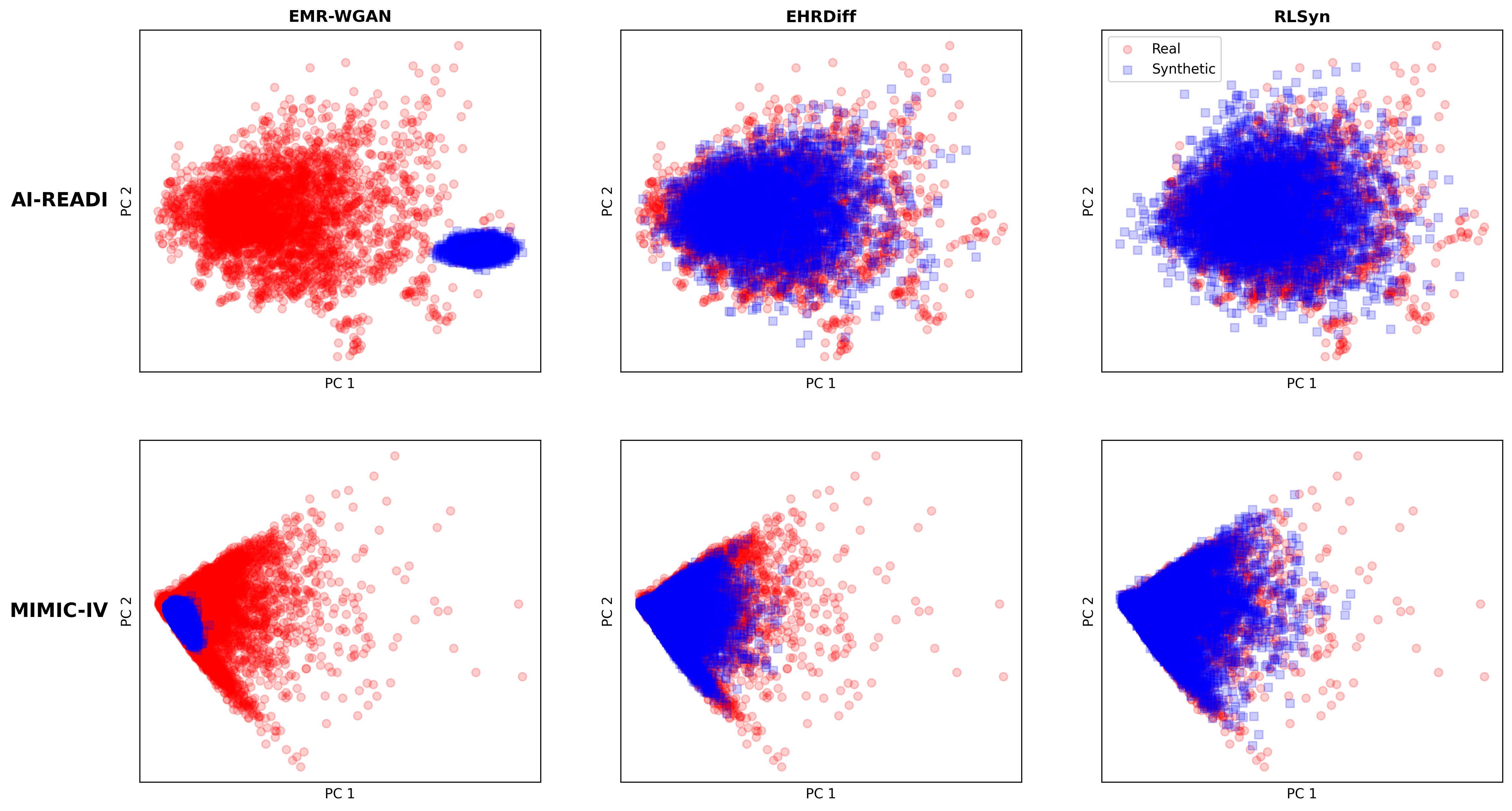}
\caption{A scatterplot of the first two principal components of the real (red) and synthetic (blue) data. }
\end{figure}

Additionally, for dimension-wise results \modelname\ outperforms both EMR-WGAN and EHRDiff with the lowest DWD score (Table 2). Figure 4A indicates the tendency for EMR-WGAN to undergenerate many of the less-common conditions in both MIMIC-IV and AI-READI, while overgenerating many of the more common conditions. 
By contrast, EHRDiff and \modelname\ generate synthetic data that very nearly match the original prevalence patterns with \modelname\ slightly outperforming EHRDiff (lower APD, Figure 4A). Similarly, the distributions of continuous variables for EMR-WGAN are narrow, failing to cover the full distribution of the real data while EHRDiff and \modelname\ match these distributions well (Figure 4B).

\begin{figure}[!htb]
\centering
\includegraphics[width=1.0\textwidth]{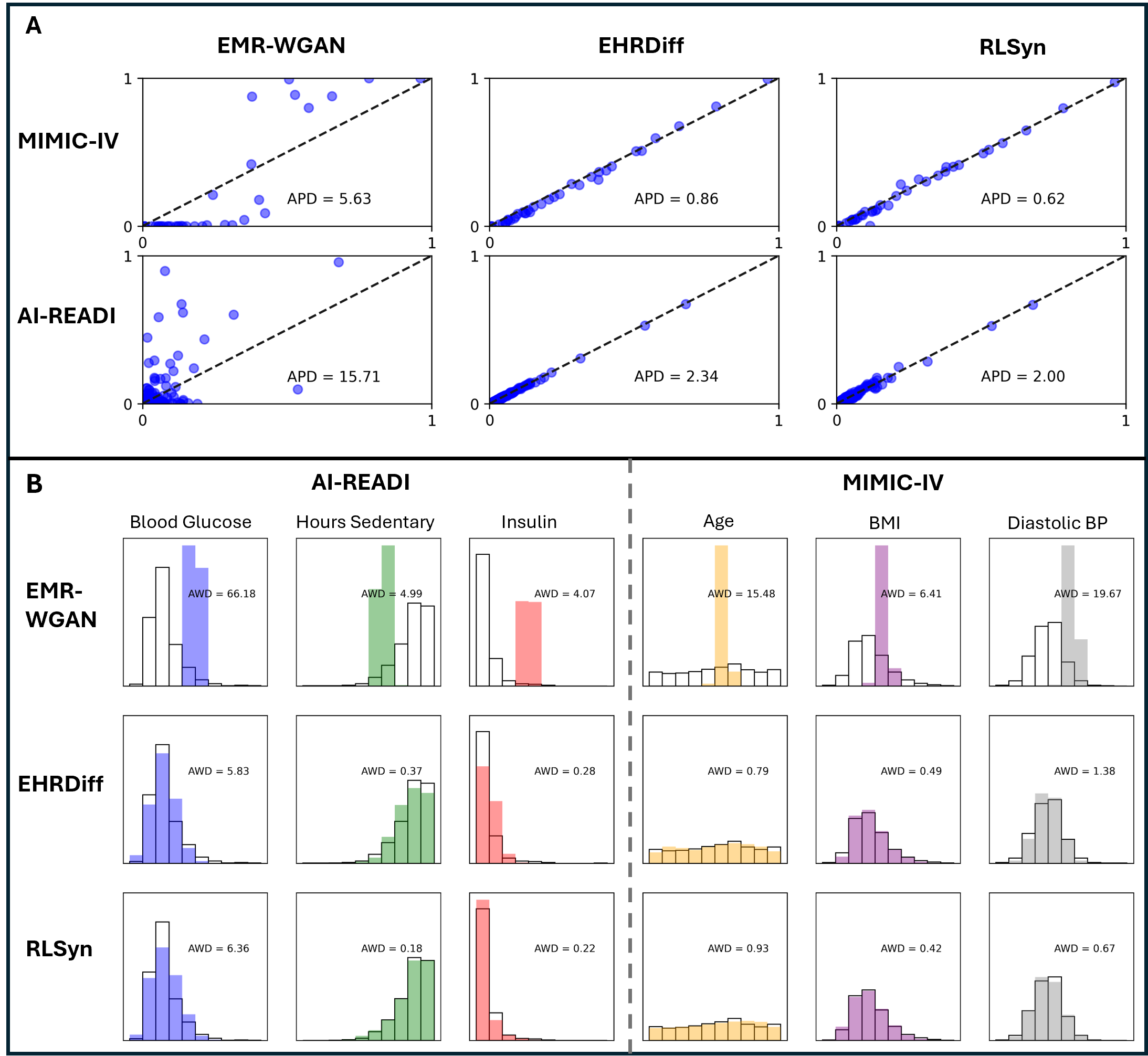}
\caption{A scatterplot of the first two principle components for real (red) vs synthetic (blue) data. }
\end{figure}
    
    \section{Discussion}\label{sec:discussion}

Our findings indicate that EHRDiff and \modelname\ outperformed EMR-WGAN in terms of utility and fidelity on both the MIMIC-IV and AI-READI datasets. Performance between \modelname\ and EHRDiff was more competitive. On MIMIC-IV, \modelname\ and EHRDiff achieved comparable utility and privacy, while \modelname\ outperformed EHRDiff in terms of fidelity. However, on AI-READI, while \modelname\ and EHRDiff still achieved comparable utility, \modelname\ outperformed EHRDiff in terms of fidelity and achieved this performance with greater privacy than EHRDiff. These results suggest that \modelname\ is more data-efficient than both GANs and diffusion approaches, producing higher-quality synthetic data with minimal privacy risks.

Taking a closer look at differences between \modelname\ and EHRDiff on the AI-READI dataset, we note that EHRDiff performed better on one fidelity metric; it better captured recreated the correlations present in the real data, evidenced by a lower CWC on both datasets. However, it had a R2S classification performance that was higher than the R2R baseline. Because the synthetic data in R2S is derived from the real training data that was used to train the classifier (unlike in S2R), performance above the R2R baseline suggests that EHRDiff overfit. This likely led to EHRDiff memorizing this training data and then recreating close proxies of the real training data. This is further evidenced by EHRDiff's higher susceptibility to membership inference attacks. Additionally, PCA visualizations show that EHRDiff’s synthetic data for AI-READI fails to cover the full data distribution captured by \modelname. Collectively, these results indicate that \modelname\ more completely captures the diversity of the data and does so with lower privacy risks than EHRDiff. This success on the smaller AI-READI dataset suggests that \modelname\ may be particularly useful in small-sample settings. 

With respect to EMR-WGAN, we note that it achieved the lowest vulnerability to membership inference attacks on AI-READI. However, this apparent privacy advantage is likely a consequence of EMR-WGAN failing to adequately learn the underlying data distribution, as reflected by its substantially lower utility and fidelity scores on both the AI-READI and MIMIC-IV datasets.

We believe there are two primary reasons 
for \modelname's effectiveness. First, in comparison to GANs, \modelname\ invokes a more principled objective for generative modeling than adversarial training. In particular, GANs optimize a nonstationary objective defined by the discriminator, which makes training unstable and highly sensitive to hyperparameter selection. For example, a common mode of failure for GANs is mode collapse, where the generator learns to generate the most common data point, which often fools the discriminator. Conversely, when the discriminator is too powerful, it may fail to provide meaningful feedback to the generator, often referred to as the vanishing gradients problem \cite{arjovsky_wasserstein_2017}. \modelname\ mitigates these issues via the use of PPO, which constrains generator updates through its clipped objective. This ensures that the distribution learned at each training iteration remains similar to that of the previous iteration, avoiding the instability of GAN training. 

The second likely reason for \modelname's effectiveness is its data-efficiency. Specifically, while diffusion models are powerful, they require deep networks and long sampling trajectories. This may contribute to their tendency to memorize training data \cite{carlini_extracting_2023, rahman_investigating_2025}. We observed that EHRDiff overfit the smaller AI-READI dataset, resulting in the elevated membership-inference risk. By contrast, \modelname\ does not have this issue because PPO optimizes an expected reward under stochastic policies, allowing \modelname\ to learn a distribution over outputs rather than collapsing onto high-probability training points (i.e., memorization). This provides two potential benefits. First, it may act as a soft privacy mechanism, making it more difficult for the generator to memorize and replicate real data points. Second, by explicitly learning both a mean and variance parameter for each distribution, it encourages the generation of diverse samples, which likely improves the downstream predictive utility of the synthetic data. Together, these factors may explain why \modelname\ exhibits higher fidelity, particularly for the AI-READI dataset, without exhibiting the privacy vulnerabilities observed with EHRDiff. These attributes highlight the potential for \modelname\ to generate high-quality synthetic data from small-sample medical datasets.

Despite the merits of this investigation, there are several limitations of this work that we wish to highlight. The first is that our experiments focused on cross-sectional tabular data generation. This was done because many biomedical datasets are tabular (e.g., those derived from electronic health records) and is an appropriate starting point for RL-based SDG. However, SDG has been successful across many different types of medical data, including longitudinal tabular data \cite{liang_generating_2025}, clinical narrative text \cite{baumel_controllable_2024, li_are_2021} and medical imaging \cite{khosravi_synthetically_2024, ktena_generative_2024}. As a result, future work should explore the efficacy of \modelname\ for these higher-dimensional generation problems. The second limitation is that the mechanisms by which \modelname\ outperforms other methods are not yet clear. We have suggested that this is due to the efficiency of the PPO algorithm or that the reparameterization trick enables us to explicitly sample from the generative model, yielding increased diversity with lower data replication. However, we cannot conclusively claim which components of \modelname\ are most influential. Additionally, there are theoretical connections between GANs, energy-based models, and inverse RL (which is the paradigm most analogous to \modelname) \cite{finn_connection_2016}. Consequently, future work should investigate the theoretical underpinnings of \modelname\ to better understand its efficacy and attribute which specific components drive its gains. Finally, we note that an advantage of framing synthetic data generation as a reinforcement learning problem is the flexibility of the reward signal. Future work could explore alternative reward formulations—such as those based on statistical or distributional metrics—that eliminate the need to train a discriminator altogether or more directly encode objectives such as privacy or utility.

    \section{Conclusion}\label{sec:conclusions}

In this paper, we introduced and evaluated \modelname\, a novel approach to synthetic data generation that leverages reinforcement learning to create high-quality synthetic biomedical datasets. Our experiments indicate that \modelname\ performs similarly to current state-of-the-art generative models on large datasets, but its superior efficiency allows \modelname\ to outperform other models for smaller datasets, while preserving patient privacy. This novel method is a promising approach to generating high-quality, privacy-preserving datasets to facilitate biomedical research.


    \clearpage

    \bibliography{mybibfile}

    \clearpage

    \section{Acknowledgments}\label{sec:acknowledgments}

This work was supported by the National Institutes of Health (OT2OD032644, T15LM007450, U54HG012510) and the National Science Foundation REU Site Award \#2348793.

    \section{Declaration of Competing Interests}\label{sec:declaration}

The authors have no conflicts of interest to declare.

    \appendix

\section{Selected Hyperparameters}\label{sec:appendix}

\begin{table}[h!]
    \centering
    \caption{Best Hyperparameters}
    \setlength{\tabcolsep}{4pt} 
    \begin{tabular}{c|ccc|ccc}
        \toprule
        \multicolumn{1}{c|}{\textbf{Dataset}} & 
        \multicolumn{3}{c|}{\textbf{MIMIC-IV}} & 
        \multicolumn{3}{c}{\textbf{AI-READI}} \\
        \cmidrule(lr){1-1} \cmidrule(lr){2-4} \cmidrule(lr){5-7}
        \textbf{Model} & \textbf{\modelname} & \textbf{EMR-WGAN} & \textbf{EHRDiff} & \textbf{\modelname} & \textbf{EMR-WGAN} & \textbf{EHRDiff} \\
        \midrule
        Batch Size & 3840 & 512 & 2304 & 384 & 1024 & 384 \\
        Noise Dim & 128 & 128 & N/A & 128 & 112 & N/A \\
        Gradient Penalty & 5 & 10 & N/A & 5 & 5 & N/A \\
        LR$_{gen}$ & 0.0005 & 0.00001 & 0.000625 & 0.0002 & 0.00001 & 0.000075 \\
        LR$_{disc}$ & 0.0001 & 0.00002 & N/A & 0.00005 & 0.000005 & N/A \\
        Width$_{gen}$ & 128 & 384 & 448 & 128 & 256 & 384 \\
        Width$_{disc}$ & 256 & 384 & N/A & 128 & 512 & N/A \\
        Steps$_{disc}$ & 3 & 1 & N/A & 5 & 5 & N/A \\
        Layers$_{gen}$ & 1 & 5 & 3 & 1 & 3 & 3 \\
        Layers$_{disc}$ & 1 & 5 & N/A & 1 & 4 & N/A \\
        Time Dimension & N/A & N/A & 448 & N/A & N/A & 128 \\
        Epochs & 30,000 & 30,000 & 5,000 & 30,000 & 30,000 & 2,500 \\
        PPO Epochs & 2 & N/A & N/A & 3 & N/A & N/A \\
        PPO Mean Penalty & 0.0 & N/A & N/A & 0.2 & N/A & N/A \\
        PPO Clip & 0.1 & N/A & N/A & 0.1 & N/A & N/A \\
        PPO Entropy Coef. & 0.001 & N/A & N/A & 0.001 & N/A & N/A \\
        PPO Value Coef. & 0.5 & N/A & N/A & 0.5 & N/A & N/A \\
        \bottomrule
    \end{tabular}
    \begin{tablenotes}[flushleft]
    \footnotesize
    \item \textit{Note.} No tuning was performed for EMR-WGAN on MIMIC-IV; we used the hyperparameters and code from \cite{yan_generating_2024}. For \modelname, PPO clip, PPO entropy weight, and PPO value weight were not tuned, as these follow standard PPO defaults.
    \end{tablenotes}
    \label{tab:grouped}
\end{table}

\section{AI-READI Features}\label{sec:appendix}

\begin{table}[ht]
\centering
\small
\caption{Complete list of AI-READI features used (108 total).}
\label{tab:aireadifeature_list}
\setlength{\tabcolsep}{6pt}
\begin{tabular}{p{0.33\linewidth} p{0.33\linewidth} p{0.33\linewidth}}
\toprule
\multicolumn{3}{l}{\textbf{Wearable and glucose monitor-derived continuous features}} \\
\midrule
Mean Heart Rate & Mean Blood Glucose & Mean Respiration Rate \\
Mean Stress & Std Blood Glucose & Resting Heart Rate \\
Total Steps & Total Calories Burned & Light Sleep Hours \\
Deep Sleep Hours & REM Sleep Hours & Awake Hours \\
Generic Activity Hours & Walking Activity Hours  & Running Activity Hours \\
Sedentary Activity Hours & & \\
\midrule
\multicolumn{3}{l}{\textbf{Clinical, cognitive, and ECG continuous features}} \\
\midrule
A/G Ratio vaƒlue & ALT (IU/L) & AST (IU/L) \\
Albumin (g/dL) & Alkaline Phosphatase (IU/L) & BUN (mg/dL) \\
BUN/Creatinine ratio & Bilirubin Total (mg/dL) & C-Peptide (ng/mL) \\
CRP--HS (mg/L) & Calcium (mg/dL) & Carbon Dioxide Total (mEq/L) \\
Chloride (mEq/L) & Creatinine (mg/dL) & Globulin Total (g/dL) \\
Glucose (mg/dL) & HDL Cholesterol (mg/dL) & Insulin (ng/mL) \\
LDL Cholesterol (mg/dL) & Potassium (mEq/L) & Protein Total (g/dL) \\
Sodium (mEq/L) & Total Cholesterol (mg/dL) & Triglycerides (mg/dL) \\
Urine Albumin (mg/DL) & Urine Creatinine (mg/DL) & Digit Span Time value  \\
Delayed Recall With No Clue Time value & Clock Visuospatial Executive Time value &  Cube Visuospatial Executive Time value \\
Memory Trial 1 Time value  & Memory Trial 2 Time value  & Moca Abstraction Time value \\
Moca Orientation Time value & Moca Total Score Time value & Naming Time value \\
Subtraction Time value & Repetition Time value  & Lettera Time value \\
Trails Visuospatial Executive Time value & Moca Combined Mis Score value & Memory Trial 1 value \\
Memory Trial 2 value & Moca Abstraction value & Moca Orientation value \\
Naming value & Repetition value & Subtraction value \\
PR (ECG) & QRSD (ECG) & QT (ECG)\\
QTc (ECG) & P (ECG) & QRS (ECG) \\
T (ECG) & & \\
\midrule
\multicolumn{3}{l}{\textbf{Categorical condition and diagnostic features}} \\
\midrule
Age-related macular degeneration & Arthritis & Cancer \\
Cataracts (1+ eyes) & Chronic pulmonary problems & Circulation problems \\
Diabetic retinopathy (1+) & Digestive problems & Marijuana user \\
Dry eye (1+) & Glaucoma (1+) & Hearing impairment \\
Heart attack & High blood cholesterol & High blood pressure \\
Kidney problems & Low blood pressure & Mild cognitive impairment \\
Multiple sclerosis & Obesity & Osteoporosis \\
Other heart issues (pacemaker) & Other neurological conditions & Parkinson's disease \\
Retinal vascular occlusion & Stroke & Type 2 Diabetes \\
Urinary problems & Abnormal (ECG) & Borderline (ECG) \\
Normal (ECG) & Otherwise Normal (ECG)  & Fluency Language value\\
Cube Visuospatial Executive value & Lettera value & Trails Visuospatial Executive value \\

\bottomrule
\end{tabular}
\end{table}

A complete list of all AI-READI features used in our experiments is provided in Table~\ref{tab:aireadifeature_list}.

\end{document}